\documentclass[letterpaper, 10 pt, journal, twoside]{IEEEtran}
\usepackage{amsmath,amsfonts}
\usepackage{mathtools}
\usepackage{verbatim}
\usepackage{graphicx} 
\usepackage{placeins}
\usepackage{multicol}
\usepackage{xfrac}
\usepackage{gensymb}
\usepackage{float}
\usepackage{bm}
\usepackage{gensymb}
\usepackage[binary-units,product-units=power]{siunitx}
\usepackage{graphicx}

\usepackage{mathtools}
\usepackage{amsmath}
\usepackage{amssymb}
\usepackage{amsfonts}
\usepackage[mode=buildnew]{standalone}
\usepackage{booktabs} 
\usepackage[caption=false,font=footnotesize]{subfig}
\usepackage{nccmath}

\usepackage{enumitem} 
\usepackage{cite}
\usepackage{float}
\usepackage{siunitx}
\usepackage[T1]{fontenc}
\usepackage{tikz}
\usepackage{acro}
\usepackage[thinc]{esdiff}

\usepackage{booktabs}
\usepackage{multirow}
\usepackage{makecell}
\usepackage{colortbl}
\usepackage{arydshln}
\usepackage{lipsum} 
\usepackage[flushleft]{threeparttable}

\usepackage{authblk} 
\usepackage[pdfpagemode=UseNone, hidelinks]{hyperref} 

\usepackage{amsmath,amsfonts}
\usepackage{mathtools}
\usepackage{verbatim}
\usepackage{graphicx} 
\usepackage{placeins}
\usepackage{multicol}
\usepackage{xfrac}

\DeclareCollectionInstance{plainmath}{xfrac}{mathdefault}{math}
{
scale-factor = 4.0,
scaling = true
}

\usepackage{enumitem} 
\usepackage{cite}
\usepackage{float}
\usepackage{siunitx}
\usepackage[T1]{fontenc}
\usepackage{tikz}
\usepackage[thinc]{esdiff}

\usepackage{booktabs}
\usepackage{multirow}
\usepackage{makecell}
\usepackage{colortbl}
\usepackage{arydshln}
\usepackage{lipsum} 
\usepackage[flushleft]{threeparttable}

\usepackage{authblk} 
\usepackage[pdfpagemode=UseNone]{hyperref} 

\NewDocumentCommand{\todo}{o m}{\textcolor{red}{\textbf{TODO\IfNoValueTF{#1}{}{(#1)}:} #2}}
\DeclareAcronym{DOF}{
  short = DOF,
  long  = degrees of freedom
}
\DeclareAcronym{GNSS}{
  short = GNSS,
  long  = global navigation satellite system
}

\DeclareAcronym{IMU}{
  short = IMU,
  long  = Inertial Measurement Unit
}
\DeclareAcronym{LiDAR}{
  short = LiDAR,
  long  = Light Detection and Ranging
}
\DeclareAcronym{UXO}{
  short = UXO,
  long  = unexploded ordnance
}

\DeclareAcronym{EKF}{
  short = EKF,
  long  = Extended Kalman Filter
}
\DeclareAcronym{iEKF}{
  short = iEKF,
  long  = iterated Extended Kalman Filter
}

\DeclareAcronym{LIO}{
  short = LIO,
  long  = LiDAR-Inertial Odometry
}

\DeclareAcronym{LO}{
  short = LO,
  long  = LiDAR Odometry
}

\DeclareAcronym{MAV}{
  short = MAV,
  long  = micro aerial vehicle,
  short-indefinite  = an
}

\DeclareAcronym{FOV}{
  short = FOV,
  long  = field of view
}

\DeclareAcronym{SDF}{
  short = SDF,
  long  = signed distance field,
  short-indefinite  = an
}

\DeclareAcronym{NCCR}{
  short = NCCR,
  long  = National Center of Competence in Research,
  short-indefinite = an,
  long-indefinite = a
}

\DeclareAcronym{SNR}{
  short = SNR,
  long  = Signal-to-Noise Ratio,
}

\DeclareAcronym{VIS}{
  short = VIS,
  long  = Vision Assistance Race,
}

\DeclareAcronym{OCR}{
  short = OCR,
  long  = Optical Character Recognition,
}

\DeclareAcronym{VLM}{
  short = VLM,
  long  = Vision Language Models,
}

\DeclareAcronym{VIO}{
  short = VIO,
  long  = Visual Inertial Odometry,
}

\hypersetup{
    pdftoolbar=true,        
    pdfmenubar=true,        
    pdffitwindow=false,     
    pdfstartview={FitH},    
    pdftitle={Sight Guide: A wearable assistive perception and navigation system for the Vision Assistance Race in the Cybathlon 2024},    
    pdfauthor={Patrick Pfreundschuh},     
    pdfsubject={IEEE RA Magazine},   
    pdfcreator={Patrick Pfreundschuh},   
    pdfproducer={Patrick Pfreundschuh}, 
    colorlinks=true,
    linkcolor=blue,
    linkbordercolor=white,
    filecolor=magenta,
    citecolor=red,
    urlcolor=cyan
    }
\usetikzlibrary{calc}

\usepackage{enumitem,amssymb}
\newlist{todolist}{itemize*}{2}
\setlist[todolist]{label=$\square$}
\usepackage{pifont}

\IEEEoverridecommandlockouts                              

\def\BibTeX{{\rm B\kern-.05em{\sc i\kern-.025em b}\kern-.08em
    T\kern-.1667em\lower.7ex\hbox{E}\kern-.125emX}}
\usepackage{balance}

\usepackage{subfiles}
\usepackage{ifthen}
\newboolean{printBibInSubfiles}
\setboolean{printBibInSubfiles}{true} 
\def\bib{\ifthenelse{\boolean{printBibInSubfiles}}
{\bibliographystyle{IEEEtran}\bibliography{references.bib}}
}

\title{\LARGE \bf
Sight Guide: A Wearable Assistive Perception and Navigation System for the Vision Assistance Race in the Cybathlon 2024
}

\author{
Patrick Pfreundschuh$^{\star 1}$,
Giovanni Cioffi$^{\star 2}$,
Cornelius von Einem$^{\star 1}$,
Alexander Wyss$^{3}$,
Hans Wernher van de Venn$^{3}$, 
Cesar Cadena$^{4}$,
Davide Scaramuzza$^{2}$,
Roland Siegwart$^{1}$, 
and Alireza Darvishy$^{3}$

\thanks{$^{\star}$The authors contributed equally to this work.}%
\thanks{$^{1}$Autonomous Systems Lab, ETH Z\"urich, CH.}%
\thanks{$^{2}$Robotics and Perception Group, University of Zurich, CH.}%
\thanks{$^{3}$School of Engineering, Zurich University of Applied Sciences, CH.}%
\thanks{$^{4}$Robotics Systems Lab, ETH Z\"urich, CH.}%
}

\begin{document}

\setboolean{printBibInSubfiles}{false}

\maketitle


\begin{abstract}
Visually impaired individuals face significant challenges navigating and interacting with unknown situations, particularly in tasks requiring spatial awareness and semantic scene understanding. To accelerate the development and evaluate the state of technologies that enable visually impaired people to solve these tasks, the \ac{VIS} at the Cybathlon 2024 competition was organized. In this work, we present Sight Guide, a wearable assistive system designed for the \ac{VIS}. The system processes data from multiple RGB and depth cameras on an embedded computer that guides the user through complex, real-world-inspired tasks using vibration signals and audio commands. Our software architecture integrates classical robotics algorithms with learning-based approaches to enable capabilities such as obstacle avoidance, object detection, optical character recognition, and touchscreen interaction. In a testing environment, Sight Guide achieved a 95.7\% task success rate, and further demonstrated its effectiveness during the Cybathlon competition. This work provides detailed insights into the system design, evaluation results, and lessons learned, and outlines directions towards a broader real-world applicability.

\end{abstract}


\section{Introduction}\label{section:introduction}
\IEEEPARstart{I}{n} 2020, approximately 43 million people worldwide were blind, with an additional 295 million suffering from moderate to severe visual impairments~\cite{iapbVisionAtlas}. Despite advancements in medical treatments~\cite{burton2021lancet}, these numbers are projected to rise by 2050~\cite{iapbVisionAtlas}. For individuals with visual impairments, the lack of visual information about their surroundings poses substantial challenges in daily activities.

While infrastructure adaptations, such as making public transport more accessible, can mitigate some difficulties, many everyday tasks remain impracticable for blind individuals. To enhance their autonomy, most visually impaired people rely on assistive technologies. 
Assistive technologies in this context are hardware- and software-based solutions that help people with disabilities to overcome or to reduce barriers in their lives. 
Although a variety of vision aids leveraging computer vision and artificial intelligence are available on the market, these solutions are typically limited to specific tasks like text-to-speech conversion~\cite{tts}, description of the surrounding \cite{chang2024worldscribe}, or navigation assistance~\cite{katzschmann2018safe, wang2017enabling}. A unified device that integrates and extends these capabilities to address a wide range of daily challenges has yet to be developed.

To advance research and innovation in this area, the 2024 Cybathlon event~\cite{bara2024cybathlon} introduced the \ac{VIS}. In this competition, blind pilots use dedicated assistive devices to complete ten tasks that reflect real-world challenges faced by visually impaired individuals. These tasks can be broadly divided into two categories: obstacle avoidance and scene understanding.

\begin{figure}[t]
    \includegraphics[width=\linewidth]{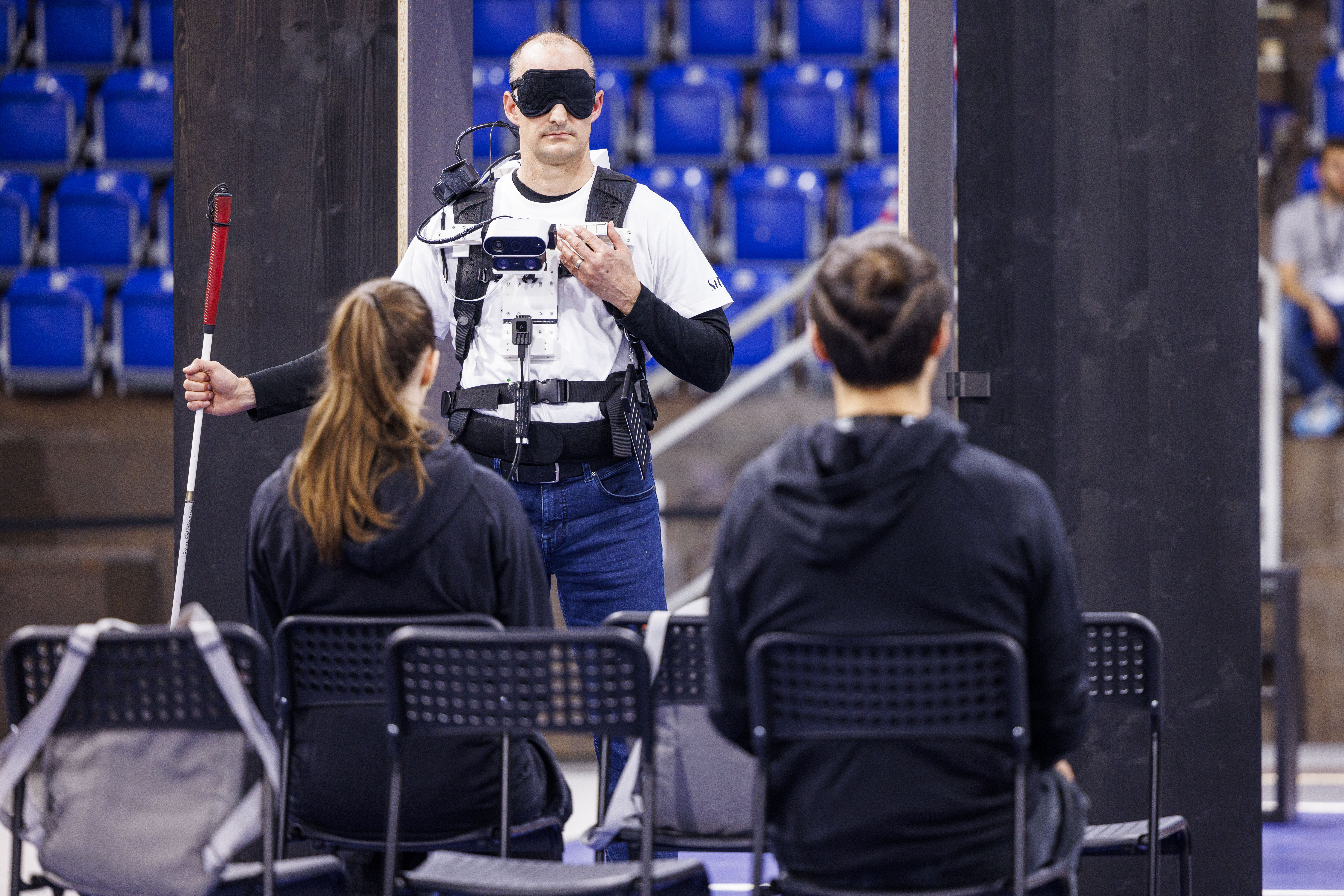}
    \caption{Sight Guide in the Cybathlon 2024 competition. The blind pilot prepares to navigate the \textit{Empty Seats} task equipped with the Sight Guide assistive device for perception and navigation. (Photo by Cybathlon / Nicola Pitaro)}
    \label{fig:intro}
\end{figure}

The obstacle avoidance tasks require pilots to navigate environments with various complexities without touching any obstacles with their white cane: maneuvering around objects like scooters, chairs, and bins (\textit{Sidewalk}); walking along a narrow path (\textit{Footpath}); and traversing a course filled with thin wooden sticks (\textit{Forest}). Scene understanding tasks test specific interactions, such as reading a name and pressing the corresponding doorbell button (\textit{Doorbell}), identifying and indicating unoccupied chairs (\textit{Free Chairs}), retrieving a specific tea box from a shelf (\textit{Grocery}), locating a target item among multiple objects on the floor without touching others (\textit{Finder}), sorting T-shirts by color and brightness (\textit{Colours}), and selecting an item from a menu on a touchscreen (\textit{Touchscreen}).

In this work, we present the technical details of the assistive device developed by Team Sight Guide for the \ac{VIS}. Our solution integrates multiple RGB and depth cameras mounted on the pilot's chest, which are connected to an embedded computer carried in a backpack. A vibration belt gives directional information to the pilot for navigation tasks, while a speaker provides instructions for scene understanding. We describe our software stack that combines classical and learned algorithms to solve each task of the \ac{VIS}. We evaluate the system in our training environment and discuss the results during the Cybathlon 2024.
\bib


\section{System Description} \label{sec:system}
We present a comprehensive overview of the system, beginning with the hardware platform and followed by the software architecture. The hardware is designed as a wearable, sensor-integrated system, while the software stack processes sensor data and generates user feedback. Together, these components enable the system to perceive the environment to provide meaningful assistance to the user in the \ac{VIS} tasks.
\subsection{Hardware}\label{sec:hardware}

\begin{figure*}
    \centering
    \includegraphics[width=\textwidth]{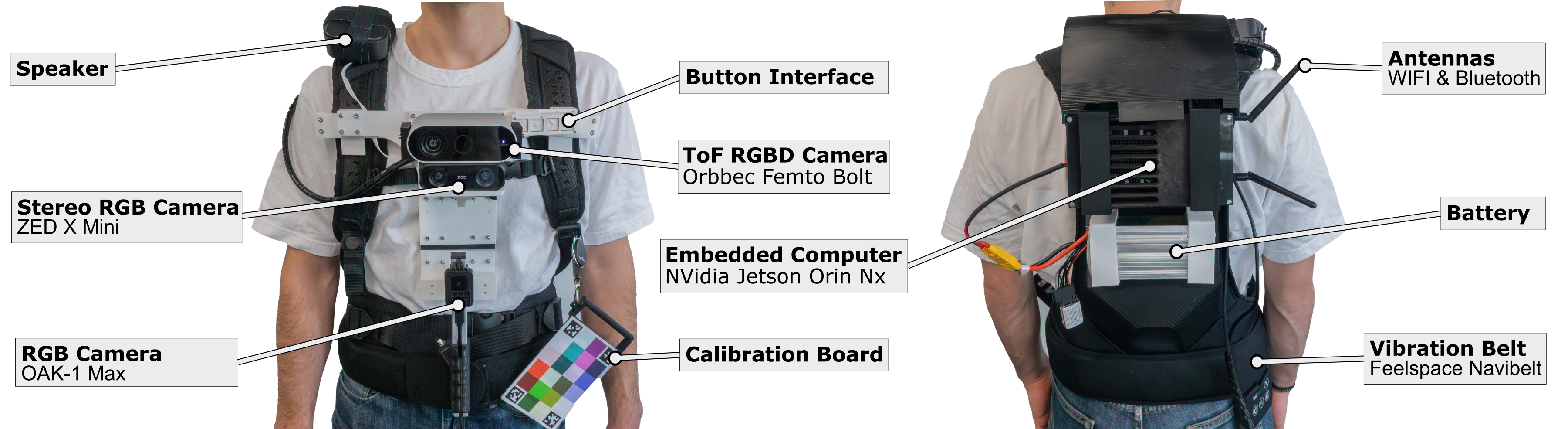}
    \caption{Hardware Overview. The wearable Sight Guide system is built around a backpack, that carries an embedded computer and the battery at the back of the pilot. The front features a multi-camera system with two cameras mounted on a 3D-printed chest plate and one handheld, moveable camera that can be attached to the chest plate when not in use. A speaker on the shoulder provides audio signals and a vibration belt provides spatial information. The user operates the device using a button interface at the chest.}
    \label{fig:hardware}
\end{figure*}
The hardware system is designed as a wearable platform that integrates multiple sensors to perceive the environment, a compact compute unit for onboard data processing, and components for delivering both audio and haptic feedback to the user. The following sections describe the individual components in detail. An overview of the system is provided in Figure~\ref{fig:hardware}.
\subsubsection{Sensors}\label{sec:sensors}
Our system integrates stereo images and IMU data from a ZEDx Mini. These sensor measurements are used to estimate the 6-DoF pose of the camera using \ac{VIO}~\cite{huang2019visual}. Although this camera also provides depth information through stereo vision, its accuracy proved insufficient to reliably detect thin objects, such as the poles in the forest task. To address this limitation, we incorporate an Orbbec Femto Bolt as an additional depth sensor. Although the Orbbec offers high-precision depth estimation on most surfaces, it struggles with objects with low reflectivity such as the black chairs in the competition due to its time-of-flight technology. To compensate for this, we fuse depth data from both cameras by filling in missing pixels in the Orbbec’s depth image with reprojected depth values from the ZED camera.
For scene detection tasks which require sharp, high-resolution images, such as \textit{Doorbell}, \textit{Grocery}, and \textit{Touchscreen}, we equip the pilot with a Luxonis OAK-1 Max. This camera was chosen because of its autofocus capability and 32MP resolution, which help to perceive small fonts or logos that are necessary for these tasks.
\subsubsection{Feedback}\label{sec:feedback}
Our device conveys information to the pilot through two modalities: vibrations and sound. Sound is commonly used by blind individuals to interact with digital devices such as smartphones and can communicate rich information. In an early prototype, we used audio cues for navigation. Although this was shown to be generally effective, it proved challenging to convey fine-grained directional information efficiently. 
To enhance spatial awareness and navigation efficiency, we integrated a Feelspace Navibelt, a vibration belt equipped with 16 vibrational units, spaced equally around the body, connected via Bluetooth Low Energy (BLE). When combined with orientation data from \ac{VIO}, the belt provides real-time directional guidance within a global frame of reference, whose origin is set using semantic information of the task layout. This method resulted in significantly improved navigation accuracy and speed compared to audio feedback.
\subsubsection{Compute}\label{sec:compute}
We process sensor data onboard using an NVIDIA Jetson Orin NX with 16GB of GPU memory. A Seeed Studio J401 carrier board connects the different sensors to the Jetson. The system runs on Ubuntu 20.04 and leverages ROS to read sensor data.

Due to the limited computational resources, running the entire software stack simultaneously is not feasible. Instead, we implemented a state machine on top of rosmon~\footnote{\url{https://github.com/xqms/rosmon}}, which enables transitions between different tasks based on input from the button interface.

\subsubsection{Mounting and User Interface}\label{sec:mounting}
As a basis for attaching both the computer and all our sensors to the pilot, we utilize an HP VR-Backpack G2. 
A custom 3D-printed mounting plate has been created to secure the Jetson, a custom power supply, and a battery on the backpack. 
In addition, the backpack has been modified to allow for the attachment of a 3D-printed chest plate, to which both the ZEDx Mini and the Orbbec Femto Bolt are rigidly attached. 
The Luxonis OAK-1 Max has been attached to a handle, allowing the pilot to comfortably and easily point it in different directions. 
The handle can also be attached to the chest plate by the pilot, to allow him to have both hands free for certain tasks, as well as to fix the camera orientation for certain tasks, such as \textit{Doorbell}. 

In addition, we have created a simple user interface for the pilot. 
Four buttons have been placed on the chest plate in addition to one button located on the handle of the OAK-1 Max camera. 
These buttons are wired to the GPIO pins of the Jetson and have been programmed to either trigger certain functions, based on the individual challenge, switch between tasks, or to repeat the last audio feedback. 
\subsection{Software}\label{sec:software}
Our software architecture consists of modular components, each designed for specific tasks such as navigation and scene understanding. It combines classical robotics algorithms with modern learning-based methods to process and interpret onboard sensor data. This hybrid framework enables the system to analyze its environment to give informative spatial and auditory feedback to the user. The following sections provide a detailed description of the individual components of the software architecture.

\begin{figure}[b]
    \vspace{-6mm}
    \includegraphics[width=\linewidth]{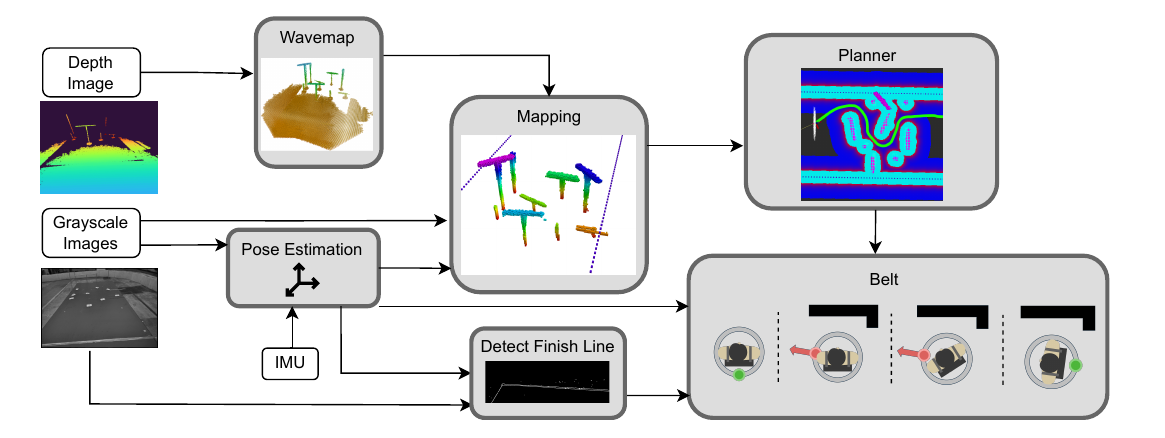}
    \caption{Overview of the navigation pipeline. The depth image is used to map occupied space, while grayscale images are used both for estimating the pilot's pose and detecting the borders of the task layout. The system computes the difference between the desired and current forward directions. Based on this difference, the belt vibrates to provide the pilot with directional feedback. In the \textit{Belt} module: From left to right: \textit{First}: The center-back belt unit (green) vibrates, indicating that the pilot can move forward. \textit{Second}: A unit on the left vibrates (red), signaling that the pilot is facing an obstacle and should rotate left. \textit{Third}: The pilot rotates and the vibrating unit on the belt changes, such that the vibrating position stays the same in the global frame. \textit{Fourth} The center-back unit vibrates again, indicating that the pilot can move forward.}
    \label{fig:navigation}
\end{figure}
\vspace{-1mm}

\subsubsection{Navigation}\label{sec:navigation}
The navigation module is responsible for estimating the pilot's pose, mapping, planning a path through the free space to reach the desired goal, and providing feedback via the belt to guide the pilot.
Fig.~\ref{fig:navigation} shows a block diagram of the navigation module.
This module serves as the core solution for the tasks of \textit{sidewalk}, \textit{footpath}, and \textit{forest}. 
Additionally, some of its components are utilized in other tasks that require spatial awareness and guidance, such as \textit{Empty Seats}.

We use the \ac{VIO} algorithm SVO PRO~\footnote{\url{https://github.com/uzh-rpg/rpg_svo_pro_open}} as the state estimator. 
SVO PRO processes images from the stereo camera along with IMU measurements to provide the camera's 6-DoF pose in real-time.
Various VIO algorithms~\cite{huang2019visual} exist in the literature, but we selected SVO PRO for its balanced trade-off between accuracy and efficiency. 
SVO PRO combines a front end~\cite{Forster17troSVO}, which estimates the camera's pose and builds a local sparse 3D map, with a back end~\cite{leutenegger2015keyframe}, which refines the camera's poses and 3D map using motion predictions derived from the IMU measurements. 
%
%
%
%

The mapping module takes camera poses and depth images as inputs. 
For mapping, we use Wavemap~\cite{reijgwart2023wavemap}, which offers state-of-the-art performance in terms of efficiency and memory usage. 
This was the primary factor in our decision to use it as the mapping component.
Wavemap achieves its desirable computational and memory characteristics by combining Haar wavelet compression with a coarse-to-fine measurement integration scheme. 
We use it to construct a map of the surroundings within a fixed Cartesian reference frame. 
The origin of this frame is initialized at the first estimated camera pose, with the z-axis aligned to gravity. 
%
%
Gravity estimation is derived from accelerometer readings, assuming a static camera during initialization.
As the camera moves, Wavemap incrementally builds a map of the occupied space.
However, we retain only the area of the map within the task boundaries and exclude points that belong to the floor.

The boundary detection algorithm begins by extracting edges from the grayscale image captured by the left camera of the ZED camera. 
Edge extraction is performed using the Canny edge detector, and the detected edges are then projected into 3D, based on the current camera pose, assuming they lie in the ground plane. 
The ground plane coordinates are obtained by fitting a plane to the point cloud from the depth camera. 
This point cloud is first transformed into the world frame using the current camera pose and the extrinsic calibration between the grayscale and depth cameras.
The boundaries, defined as the four corner coordinates in the world frame, are estimated by fitting a rectangle of known dimensions to the 3D edge points using a RANSAC-based approach. 
To validate the detected boundaries, we impose two conditions: (1) the angle between horizontal and vertical edges must be close to 90 degrees, and (2) the number of points belonging to each edge must exceed a predefined threshold. 
This algorithm is robust even when only three edges are visible. 
Typically, the two vertical edges are well detected, while either the front or back edge may not be visible due to the initial camera pose and the limited field of view. 
The boundary detection algorithm is executed during the initialization phase of the navigation stack.

The planning algorithm begins execution only after the boundaries have been detected. 
To achieve efficient path planning, the occupancy map generated by Wavemap is converted into a 2D cost map in the x-y plane. 
This cost map is represented as a 2D occupancy grid, where each cell is assigned a value between 0 (free space) and 255 (occupied). 
Once the 2D cost map is constructed, obstacle proximity is accounted for by inflating the cost around obstacles, ensuring that the planned path maintains a safe distance from them. 
For path planning, we use the A* algorithm~\cite{lavalle2006planning}, with the final goal set 1 meter beyond the center of the task end line (back boundary edge).
The feedback provided to the pilot is based on the discrepancy between the estimated camera heading and the desired heading of the planned path. 
To determine the vibration feedback (see Fig.~\ref{fig:navigation}), we compute the heading difference and activate the corresponding vibration unit.
The feedback frequency is set to 1 Hz, which was experimentally found to offer the best trade-off between walking smoothness and responsiveness in obstacle avoidance.

We use the left camera of the ZED to detect when the pilot crosses the finish line. 
After the pilot covers half the task distance, the finish line detection algorithm is activated. 
We project the finish line’s endpoints, estimated by the boundary detection algorithm, into the camera frame using the current pose and ground estimate. 
When these points are no longer visible, the pilot is near the finish line. 
We then track the forward distance, and when it exceeds a threshold, the task is considered complete, triggering vibration on all belt units. 
We use the measurement of the overall distance traveled to reject false positives.

\subsubsection{Doorbell}\label{sec:doorbell}
\begin{figure}[b]
    \vspace{-6mm}
    \includegraphics[width=\linewidth]{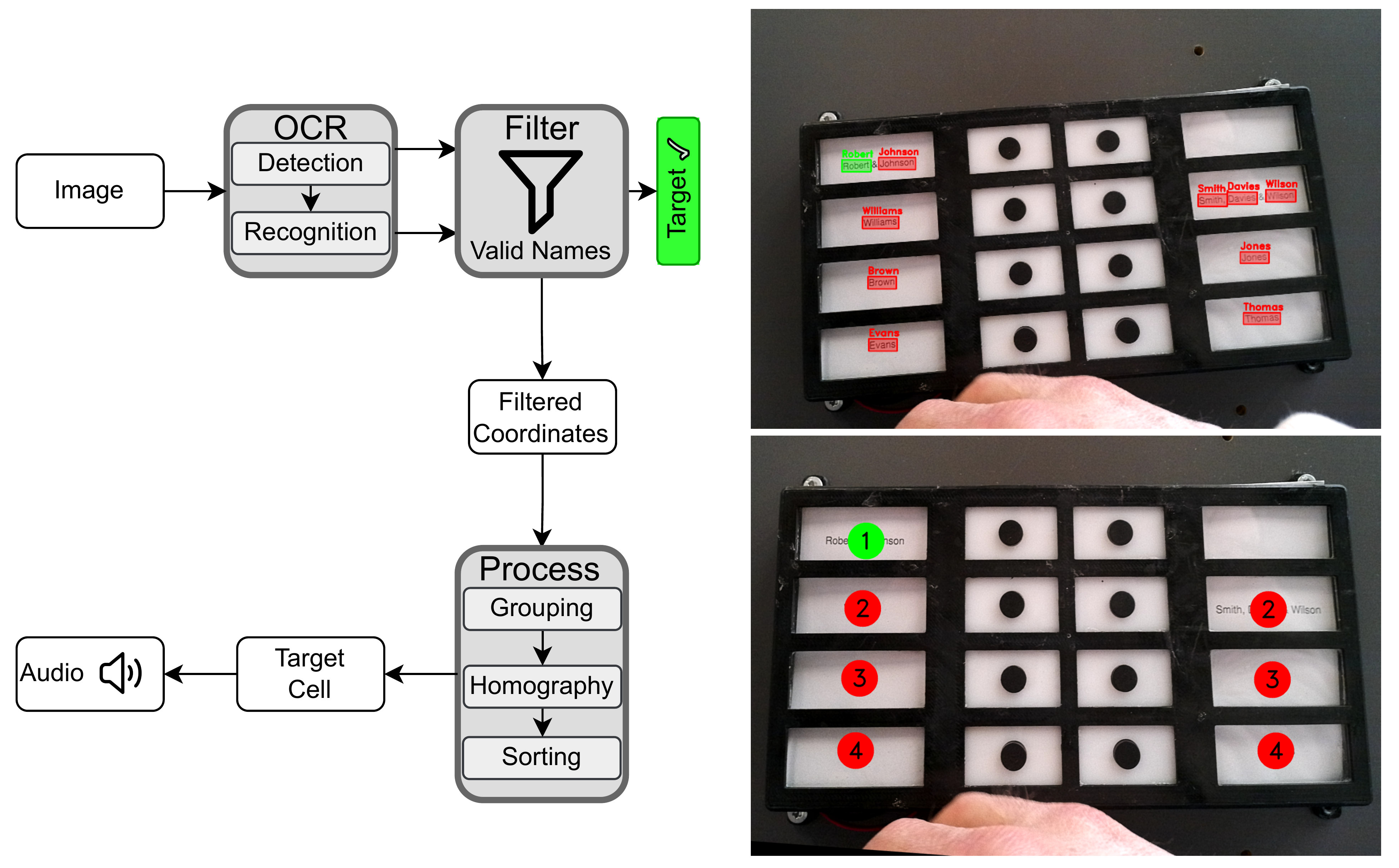}
    \caption{\textit{Left}: Overview of the doorbell task pipeline. \textit{Right Top}: Camera input image with detected names. The target name is visualized in green. \textit{Right Bottom}: Rectified image with button clusters. Numbers indicate the row index. The identified target cell is indicated in green.}
    \label{fig:doorbell}
\end{figure}

To identify text, images captured by the handheld camera are processed using an \ac{OCR} pipeline. This pipeline leverages the OnnxTR library~\footnote{\url{ https://github.com/felixdittrich92/OnnxTR}}, which provides efficient implementations of FAST~\cite{chen2021fast} for text detection and ViTSTR~\cite{atienza2021vision} for text recognition. The extracted words are compared with a list of names defined in the competition rules. To account for minor prediction errors, the matching is performed using a threshold based on the Levenshtein Distance~\cite{levenshtein1966binary}. 
At the beginning of the doorbell task, the pilot directs the handheld camera towards the table on which the card with the target name is placed. When a match is detected, the system selects the corresponding name as the target and provides an audio signal to notify the pilot. The pilot then attaches the camera to the chest plate and positions himself in front of the doorbell. Since multiple names may appear on the same nameplate in the case of shared flats, clusters are created based on the image proximity of detected text. If a sufficient number of clusters is identified, the image is rectified using the minimum bounding box around the cluster centers. Then, the rows and columns of the doorbell are determined by applying k-means clustering to the vertical and horizontal components of the name clusters in the rectified image. Finally, the row and column containing the target name are extracted and communicated to the pilot via an audio signal. A visualization of the individual steps is presented in Figure \ref{fig:doorbell}.

\vspace{-1mm}

\subsubsection{Empty Seats}\label{sec:seatfinder}
Identifying all occupied chairs from a single image is challenging due to potential occlusions caused by people sitting on the chairs. Although this issue could be solved using a specific perspective, such as a top-down view from a camera on a stick, errors in detections from a single image can also lead to incorrect classification of occupied chairs as free. To address these challenges, our system continuously constructs a semantic 3D map of the environment using RGB-D data recorded from the chest-mounted cameras as the pilot moves through the task area.

In our approach, persons, chairs, and backpacks are segmented in the RGB images using a YOLO network~\cite{khanam2024yolov11}. These segmentation masks are used to extract the corresponding 3D detections from the depth image. The classified points are then integrated into a semantic volumetric map using the Nvblox framework~\cite{millane2024nvblox}, with the camera poses provided by the \ac{VIO} pipeline (\ref{sec:navigation}). The accumulated map is analyzed once the pilot activates a respective button.

The first step involves cropping the map using the detected lines (\ref{sec:navigation}). To improve the map quality, spurious points caused by noisy depth data are removed using a statistical outlier filter. Points located between 0.4 and 1.9m above the ground are then projected onto the ground plane, and the two rows of chairs are identified as the two largest connected components. The 3D bounding box around the points inside these rows is then divided into six cells, each corresponding to a chair.

A cell is classified as occupied if the ratio of points labeled as backpack or person within it exceeds a predefined threshold. To reduce errors from points leaking into adjacent cells, only the central half of each cell is considered for evaluation. Finally, the system communicates the row and cell of each free chair to the pilot via an audio signal.

\begin{figure}[t]
    \vspace{-6mm}
    \includegraphics[width=\linewidth]{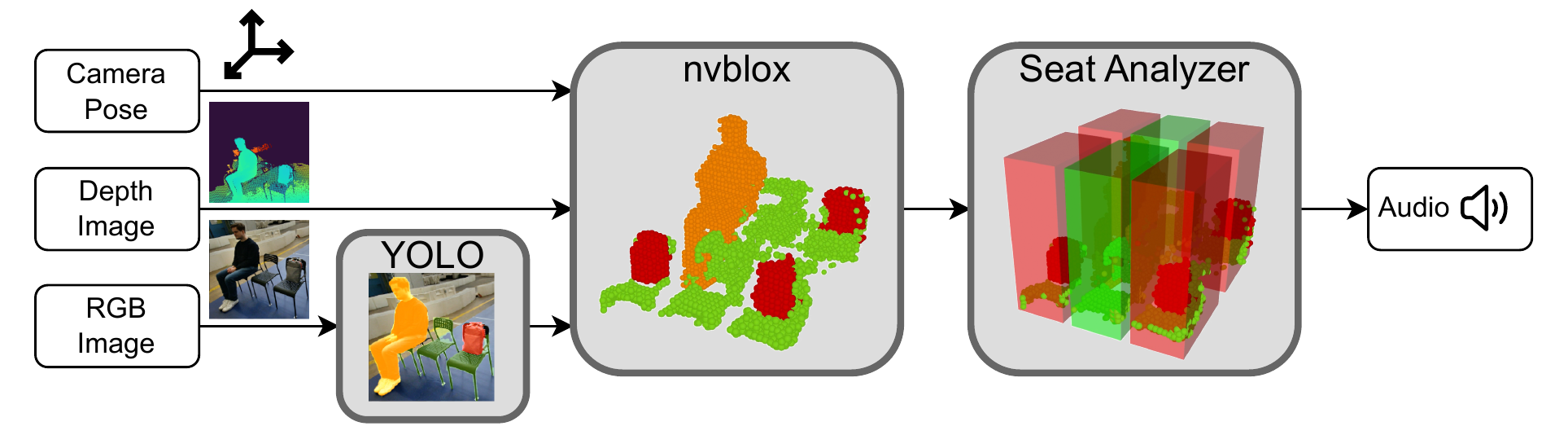}
    \caption{Overview of the \textit{Empty Seats} task pipeline. Semantic detection masks from YOLO are combined with the corresponding depth image and camera pose to build a 3D semantic map using nvblox. Upon a user input, the semantic map is analyzed to identify free seats. The rows and columns of free seats are communicated to the pilot via audio.}
    \label{fig:seats}
\end{figure}
\vspace{-1mm}

\subsubsection{Grocery}\label{sec:grocery}
\begin{figure}[t]
    \includegraphics[width=\linewidth]{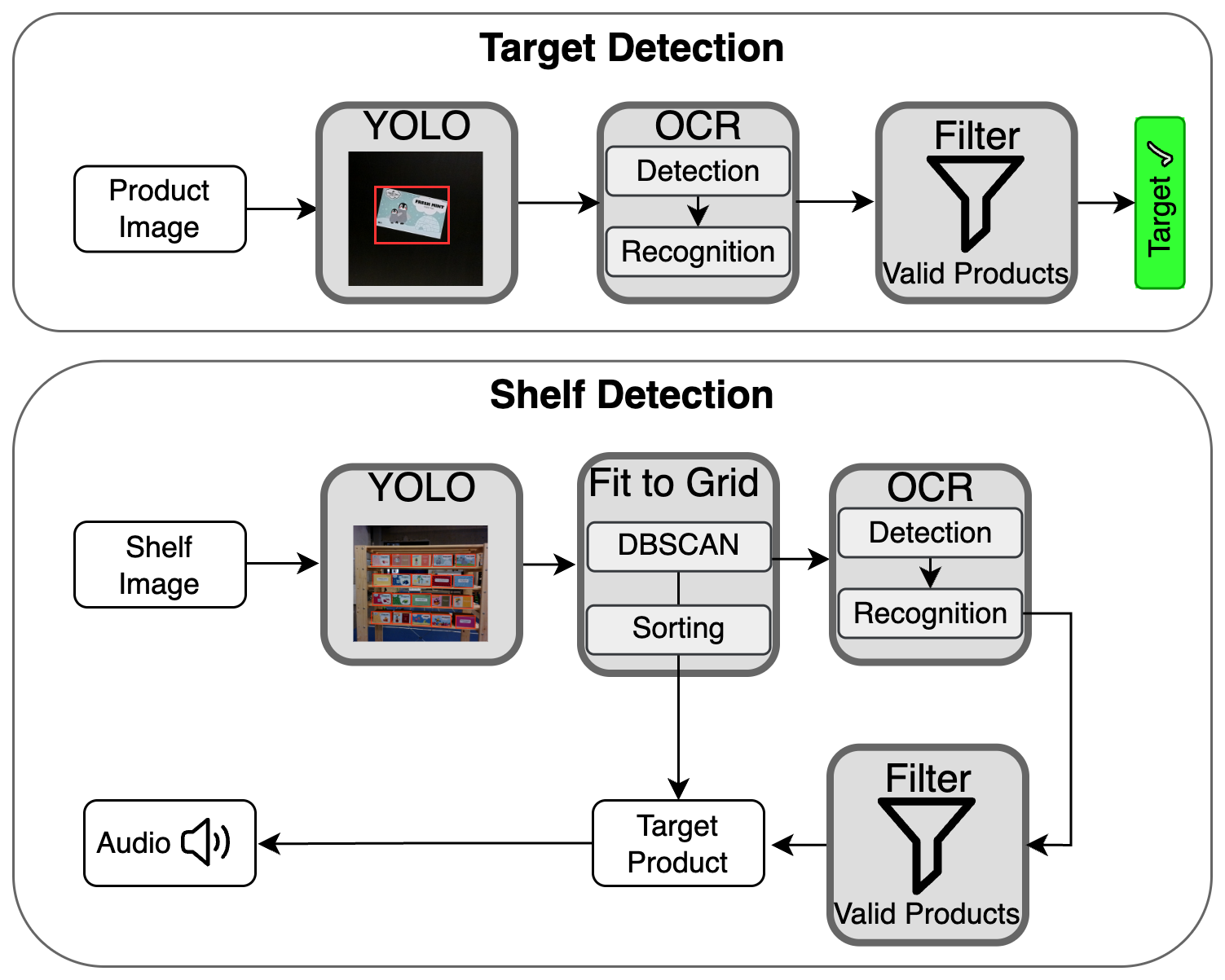}
    \caption{Grocery overview. \textit{Top:} An image of the table is taken, valid labels are detected and the target is recognized using OCR and a dictionary. \textit{Bottom:} An image of the shelf is analyzed using the same YOLO network. All products are fit to a grid and analyzed using OCR. The position of the product matching to the one on the table is communicated via audio.}
    \label{fig:grocery}
\end{figure}

In this task, the pilot starts by identifying the label of a given product on the table, and subsequently has to retrieve the matching product from a shelf containing 20 different boxes of tea. 
To visually identify both the given label and all the products on the shelf, we utilize a YOLO network ~\footnote{\url{https://github.com/ultralytics/ultralytics}} which has been fine-tuned on a custom dataset containing all the tea box labels, as defined in the competition guidelines. 
Furthermore, a dictionary containing identifying keywords has been manually created for each tea box label.
Once the given label has been detected, we utilize the same \ac{OCR} pipeline as in the \textit{doorbell} task to detect and extract words from the given label. 
These words are then compared to each keyword dictionary using a threshold based on the Levenshtein Distance~\cite{levenshtein1966binary}.
Once the source label has been matched to a keyword dictionary, it is known which object we wish to retrieve. 
The pilot now takes a second picture of the entire shelf. 
We utilize the same YOLO network to detect all products, however, to determine the row and column position of each product, wefit all detections to a grid. 
To achieve this, we apply the DBSCAN algorithm~\cite{ester1996density} on the y-coordinates of each bounding box, resulting in individual separated rows of boxes. 
Within each row, we now sort all bounding boxes by their x-coordinate, thus obtaining the row and column index of each product. 
Finally, we utilize the same \ac{OCR} pipeline to extract words from each box, which are again matched to the keyword dictionaries. 
When a match has been found, the position on the shelf is communicated to the pilot via an audio signal. In case no match is found, an audio signal requests the pilot to take a new picture and the matching process is repeated. 
This is illustrated in detail in Figure \ref{fig:grocery}.

\subsubsection{Colours}\label{sec:tshirt}
In the Colours task, the pilot has to sort 6 T-shirts that have two base colors and three levels of brightness. The two base colours are randomly chosen from a set of predefined colour combinations. The T-shirts are randomly placed on a clothes rack, and the pilot must arrange them in order of color and brightness. While it would theoretically be possible to determine the correct order from a single image of the entire rack, we use an interactive approach instead.

When the pilot reaches the clothes rack, all T-shirts are moved to the edge of the hanger. The pilot then picks up one T-shirt, holds it in front of the camera, and presses a button on the handheld camera handle. The system then determines and communicates the correct position for that T-shirt with respect to the currently sorted shirts.

A major challenge with RGB cameras is maintaining a consistent white balance. Since the camera’s automatic white balance can be affected by the environment, we use a calibration board that the pilot carries. This board has four AprilTags at its corners and includes color samples for each of the predefined colours. At the start of the task, the pilot holds the calibration board in front of the camera. The system detects the AprilTags, and uses the image area between them as a reference for white balance and exposure adjustments. The camera settings are adjusted for three seconds and then fixed for the rest of the task.

Next, the pilot holds each T-shirt in front of the camera one by one. To determine the T-shirt's base color and brightness, the system first segments the T-shirt in the image. Our segmentation process assumes that the T-shirt is the largest single-colored area in the image. We segment the image based on subsequent masking. First, we blur the image and create a histogram of hue values. We then mask the pixels that fall into the most prominent bin in the histogram. We repeat this process on the masked pixels for lightness and saturation. After that, we identify the largest connected component in the remaining masked area. The T-shirt’s color and brightness are determined based on the hue and lightness values of the average RGB color within this connected region.

\begin{figure}[t]
    \includegraphics[width=\linewidth]{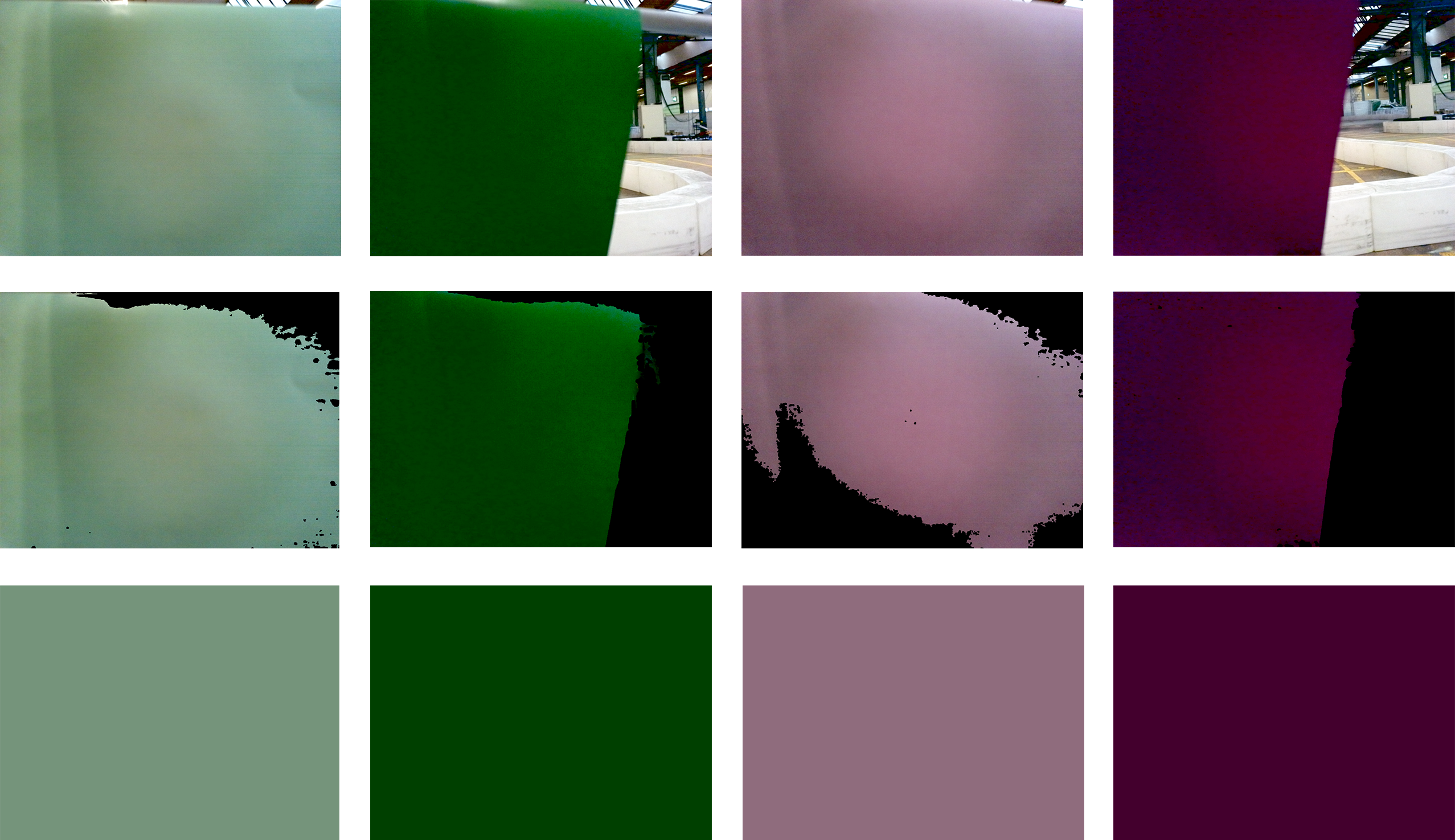}
    \caption{Colour sorting example images: The top row shows the input images to the pipeline. In the middle row, the identified colour pixels are shown in colour, while background is masked out in black. The bottom row visualizes the calculated average colour. Columns are sorted according to the two base-colours and brightness identified by the algorithm.}
    \label{fig:tshirts}
\end{figure}

Once the color and brightness are identified, the system determines the correct position for the T-shirt relative to the already sorted ones. It then calculates the most efficient way to arrange the existing shirts to minimize movement. The system provides placement instructions in the format “left | 2,” which tells the pilot to place the T-shirt in position 2, counting from the left side. This process is illustrated in Figure \ref{fig:tshirts}.

\subsubsection{Finder}\label{sec:finder}
\begin{figure}[t]
    \includegraphics[width=\linewidth]{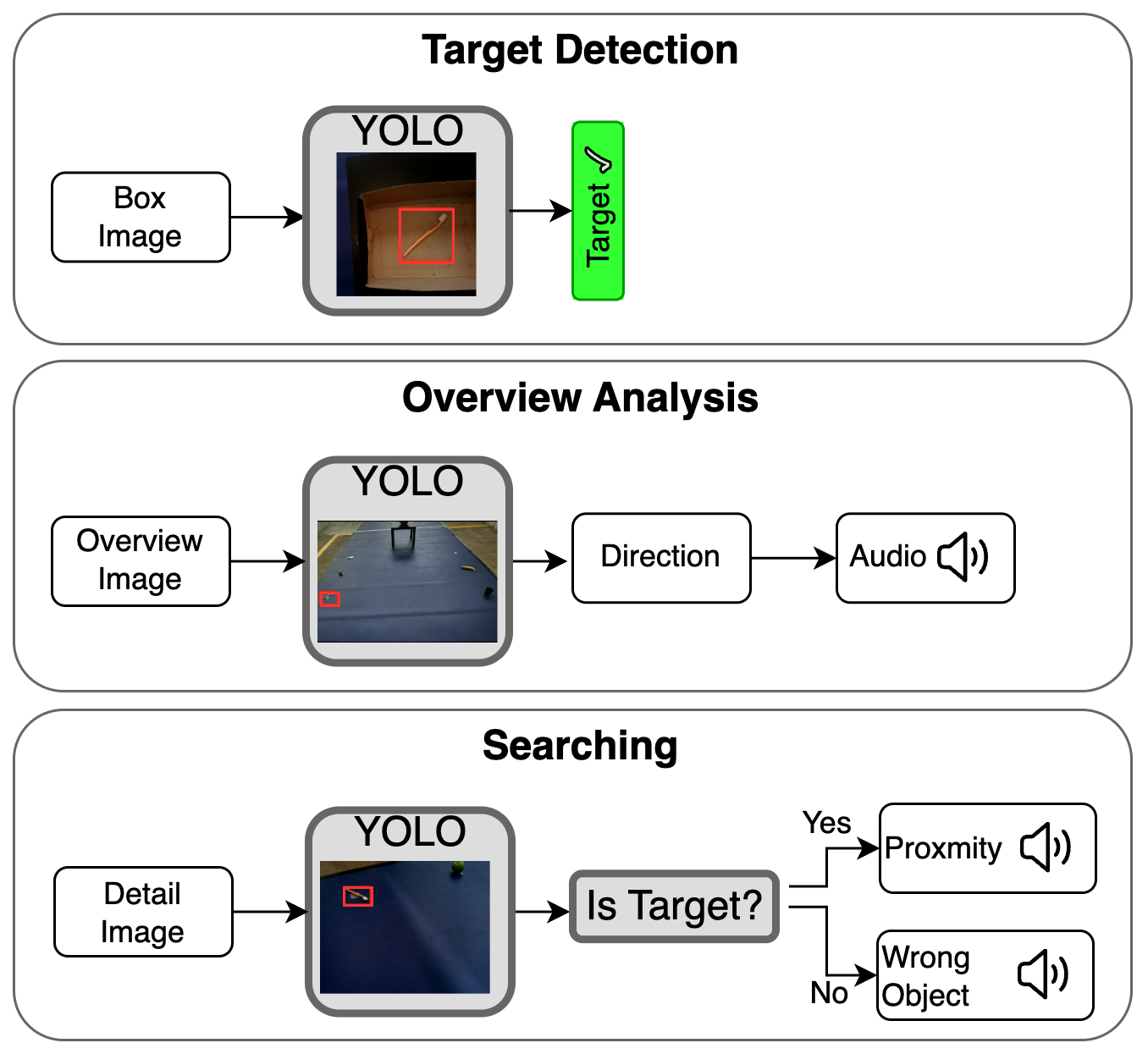}
    \caption{Finder overview. \textit{Top:} The box is opened and using YOLO the target object is determined \textit{Center:} An overview image is taken from which the rough direction of the target can be determined. \textit{Bottom:} The pilot approaches the target area and searches for the object. If the target is within reach, a proximitiy audio signal is played.}
    \label{fig:finder}
\end{figure}

In this task various objects are spread out across the floor. 
The goal for the pilot is to identify an object located in a box at the beginning of the task, find the matching object on the floor and return it to a bowl at the end of the task. 
The basis of our method is again a YOLO object detection network. 
All objects are already detectable by the standard YOLO v8 network, but we have fine-tuned this further with a custom dataset to increase the detection reliability. 
At the beginning of the task, the pilot opens the box and takes a picture with the handheld camera to identify the target object. 
Subsequently, we utilize an image from the chest mounted Orbec Femto camera to obtain an overview of the competition floor thanks to its wide-angle image. 
This can be used to guide the pilot in the rough direction of the target object, by specifying whether it is located on the left or right side of the floor. 
To precisely locate the target object, the pilot can now utilize the handheld camera and point it in various directions. 
If a non-target object is detected, it will be announced via the audio feedback to inform the pilot. 
If the target object is detected, an adaptive audio signal will be played indicating how close the pilot is to the direction of the target object. 
This process is illustrated in detail in Figure \ref{fig:finder}.

\subsubsection{Touchscreen}\label{sec:tablet}
The tablet task requires the pilot to select a specific item from a 5×5 grid of randomly arranged objects on a touchscreen. We split this task into three subproblems: Detecting the small target item on the screen, Detecting the pilot's finger and guiding the pilot to the correct screen location.

At the start of the task, the pilot positions himself roughly centered in front of the tablet. Since the target item is only 25x15mm in size, we use the 4K RGB camera mounted on the pilot’s chest to capture images of the screen. To detect the target item, we use a template-matching approach, where an image of the item serves as a reference.

As classical feature-matching methods such as SIFT and SURF, as well as learned sparse approaches like SuperPoint did not yield satisfactory results, we opted for the dense, learning-based Efficient LoFTR approach ~\cite{wang2024efficient}. Since running Efficient LoFTR on the full 4K image is computationally intractable on the onboard computer, we first use YOLO ~\cite{khanam2024yolov11} to segment the tablet screen and then apply feature matching only to the cropped screen image. We refine the matches using RANSAC, keeping only valid inliers. When the number of inliers exceeds a predefined threshold, we determine the target’s coordinates by averaging the positions of the inlier points.

Due to the movement of the pilot from breathing and moving the finger, using the camera image directly for finger navigation is impractical. Instead, we transform the image into a screen-aligned, rectified coordinate system using a homography transformation based on the detected screen corners. We refer to this rectified image as the screen image. Once the target is identified in the screen image, its position remains fixed, even if the pilot later obstructs part of the screen. This is particularly beneficial because Efficient LoFTR is too slow for continuous execution. Similarly, detecting the screen corners in every frame is impractical since the pilot’s hand may block some of them. To address this, we extract ORB ~\cite{rublee2011orb} features from the screen image and track them across subsequent frames to maintain the homography transformation.

To detect the pilot’s fingertip, we use YOLO to segment the hand and identify the fingertip as the highest detected pixel in the screen image. The pilot starts in the bottom-right corner of the screen and follows an audio-guided navigation system.

The pilot first moves his finger towards the left until reaching the target x-coordinate, with an adaptive audio signal indicating distance.
Once the correct x-position is reached, an audio cue informs the pilot to move upward.
When the fingertip is close to the target, another signal indicates that the item can be selected.
If the pilot deviates from the target coordinates, additional audio cues guide him left, right, up, or down.
The finger-tracking system operates at 5Hz.

\begin{figure}[t]
    \includegraphics[width=\linewidth]{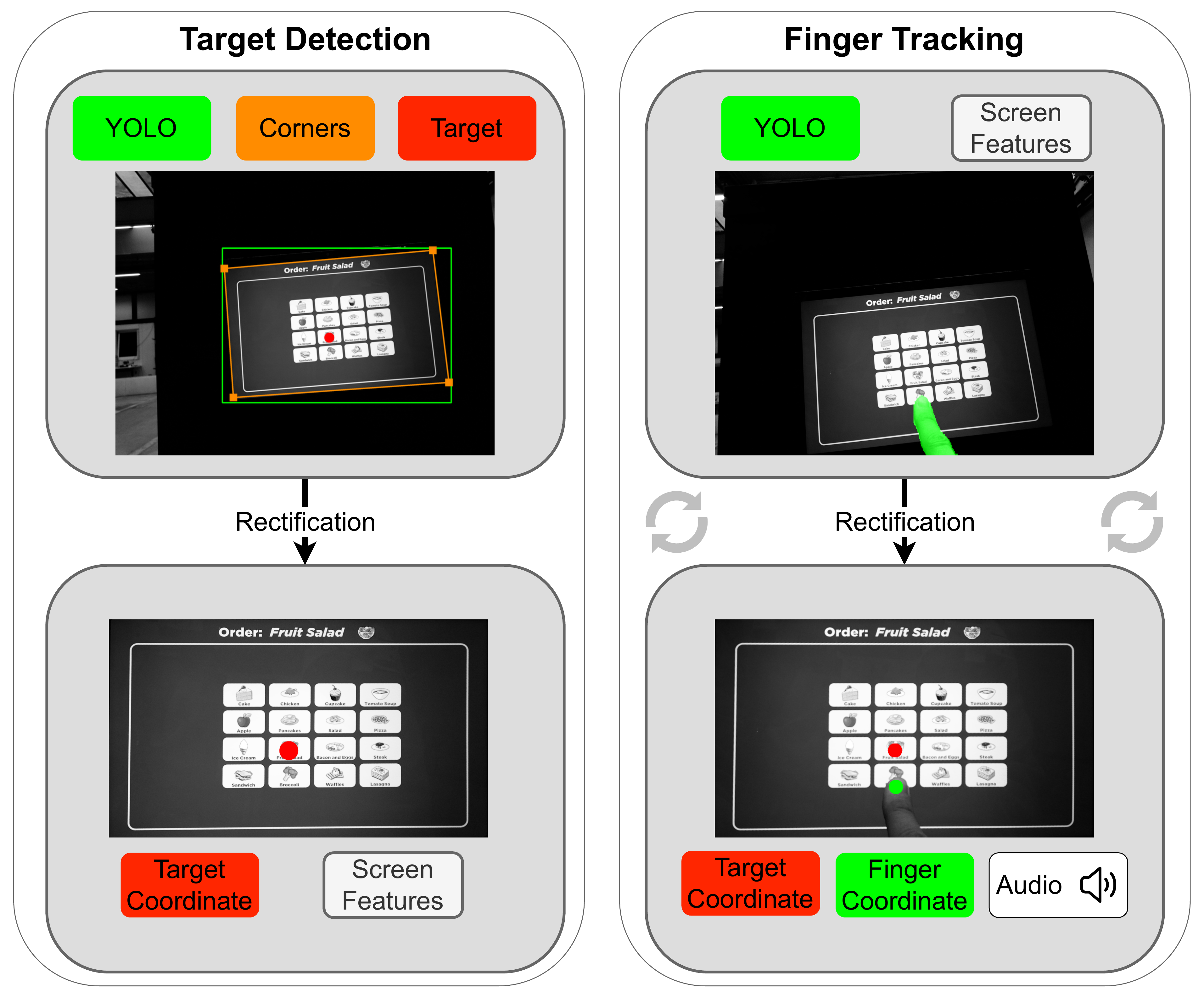}
    \caption{Screen navigation overview. \textit{Left:} First, the image is cropped using a bounding box from YOLO classification. Screen corners are found and the cropped image is matched against the target template. The image is rectified using the corner coordinates. The target coordinates and ORB features in the rectified screen frame are stored. This procedure only runs once as the target position in the screen frame is static. \textit{Right:} The finger is detected using YOLO. By matching the stored screen features with input features, the image is rectified and finger coordinate in the screen frame are found. An audio signal is generated based on finger-target distance.}
    \label{fig:screen}
\end{figure}


\section{Experimental Results} \label{section:Experiments}
We evaluate the performance of our system quantitavely in a training environment and provide qualitative insights about the results at the Cybathlon 2024 competition.

\subsection{Pilot Description and Training} \label{section:pilot}
We developed our device in an iterative process of testing and adapting to our pilot's feedback. Our pilot was born with complete blindness and was 48 years old at the time of the competition. We performed tests with him in intervals of roughly two months for 18 months. For the last month before the competition, the training frequency was increased to twice a week. While our training and development focused on one pilot, the device is not limited to him. In an additional testing session, another blind person achieved comparable performance after a 20-minute instruction to the device. This indicates the ease of use of our solution.

\subsection{Training Results} \label{section:training}
\begin{figure*}[t]
    \centering
    \includegraphics[width=\textwidth]{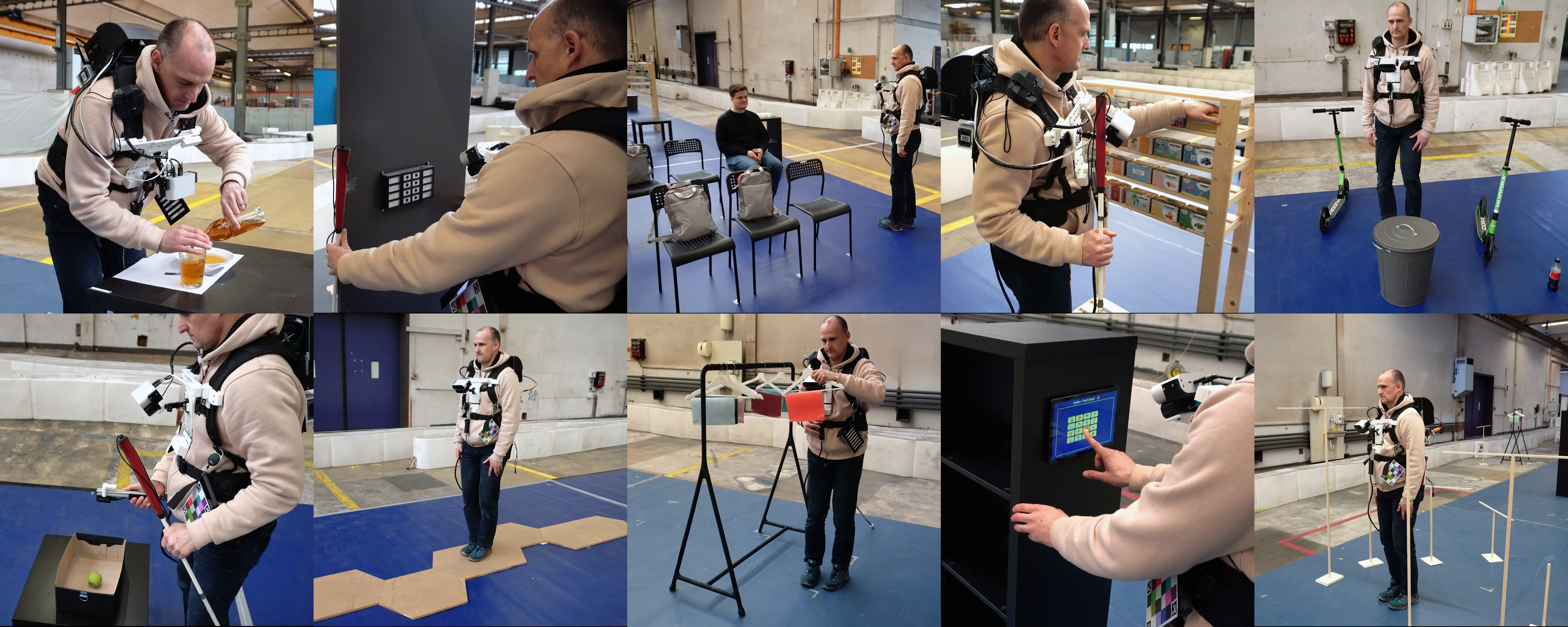}
    \caption{The blind pilot performs the Cybathlon \ac{VIS} tasks in our training environment using the Sight Guide system. Top, from left to right: \textit{Dish Up, Doorbell, Empty Seats, Grocery, Sidewalk.} Bottom, from left to right: \textit{Finder, Footpath, Colours, Touchscreen, Forest} }
    \label{fig:experiments}
\end{figure*}

To provide a realistic evaluation environment, we set up a training course according to the detailled specifications in the competition rules. Where available, we used the exact furniture and materials as in the competition. The remainder of the infrastructure was replicated according to the instructions. The setup of the individual tasks is shown in Figure \ref{fig:experiments}.

To provide comparable numbers to our competition results, we focused on the evaluation of the eight tasks we performed in the competition: \textit{Dish Up}, \textit{Doorbell}, \textit{Free Seats}, \textit{Grocery}, \textit{Sidewalk}, \textit{Colours}, \textit{Touchscreen}, \textit{Forest}. We performed ten runs, of which each run was configured using different randomizations within each task (e.g. target objects, obstacle layouts, etc). 

For each task, we analyze the success rate and time to completion. We additionally divide the success rate into device and pilot success, to distinguish errors caused by a malfunction of the device from an error caused by wrong execution of the pilot. We present our results in Table \ref{table:training}. 
For completeness, we also list results for \textit{Dish Up}, even though the pilot performed this task without any feedback from the device. 

In all runs, the device successfully solved \textit{Doorbell}, \textit{Free Seats}, \textit{Grocery}, \textit{Sidewalk}, and \textit{Touchscreen}. In one trial, the system identified the wrong target cell during the \textit{Grocery} task. Due to challenging, dark lighting conditions in our testing environment, the handheld OAK camera sometimes gave underexposed images, which resulted in a wrongly sorted result in the \textit{Coulors} task once. Additionally, the system navigated the pilot too close to an obstacle in one configuration in the \textit{Forest} task. However, averaged over all tasks (excluding \textit{Dish Up}) and runs, our system achieved a 95.7\% success rate. It can be noted, that the pilot success rate of 91.3\% is slightly lower. During two trials, the pilot accidentally put the colours in a different order than communicated by the device. Similarly, in two attempts at the \textit{Tablet} task, the device navigated the finger to the correct target item, but the pilot slightly shifted his finger while lifting it to activate the item, eventually triggering the wrong item. 

With 469 seconds, the average time for the eight tasks is below the allowed competition time of 480 seconds. The time of completion varies greatly between different tasks. Due to our vibration-based navigation, the navigation-related tasks take little time. The detection-related tasks \textit{Doorbell}, \textit{Grocery} and \textit{Tablet} take comparably long due to two factors. First, the initialization time is longer, as the required neural networks need to be loaded into the GPU memory. Second, as the pilot does not always position himself ideally in front of the target scene, it often takes some seconds until the camera focuses correctly on the target. Finally, the \textit{Colours} task requires a lot of movement sequences to arrange the hangers, resulting in a high time to completion.

Due to limited success rates or long completion times, we have decided not to attempt the \textit{Footpath} and \textit{Finder} challenges during the Cybathlon competition, even though we have developed solutions for these tasks. 
While the \textit{Finder} task could be completed with high reliability, it often took more than 114 seconds to complete. 
In the footpath task, if the pilot minimally touches the edge of the tiles, the task is considered failed. As a result, our navigation algorithm (Fig.~\ref{fig:navigation}) must guide the pilot with frequent and precise directional adjustments. This led to a low success rate. For this reason, we decided not to attempt the footpath task during the final competition.

We provide a video of a full run online~\footnote{\url{https://youtu.be/Ps3tWnj-XIg}}.

\begin{table*}[t!]
\centering
\caption{Training Results}
\begin{tabular}{l c c c c c c c c c}
\toprule
Task                    & Serving                        & Doorbell                       & Seatfinder & Grocery     & Sidewalk & Colours     & Tablet    & Forest  & All   \\
\midrule
Success Device {[}\%{]} & -                            & 100                            & 100        & 90          & 100      & 90          & 100       & 90      & 95.7    \\
Success Pilot {[}\%{]}  & 100                            & 100                            & 100        & 90          & 100      & 70          & 80        & 90      & 91.3   \\
Time {[}s{]}            & 46.2 $\pm$ 3.5 & 67.2 $\pm$ 6.7 & 27.6 $\pm$ 4.9 & 75.3 $\pm$ 20.7 & 37 $\pm$ 6.6 & 93.4 $\pm$ 12.7 & 94.9 $\pm$ 18 & 27.2 $\pm$ 2.3 &469 $\pm$ 28.4  \\
\bottomrule
\end{tabular}
\label{table:training}
\end{table*}

\subsection{Cybathlon 2024 Results} \label{section:competition}
For the actual competition runs, we targeted the 8 tasks listed in Table \ref{table:training} given the ambitious time limit of 8 minutes, and decided to skip the \textit{Footpath} and \textit{Finder} tasks, due to their longer time-to-completion. 
The competition is split into two qualification runs, followed by one final run that determines the end result. 

During the first qualification run, the device lost connection to the vibration belt. This caused a large delay and rendered several tasks unfeasible, which resulted in only four successfully solved tasks. In the second run, our system performed successfully on all targeted tasks within the allowed time limit. During the \textit{Empty Seats} task, the pilot stepped out of the allowed race track with one foot, which is counted as a failure of the task. With the remaining seven tasks solved, our team placed first in the qualification. Due to another hardware issue, this performance could not be repeated during our final run. As the Orbbec Depth Camera stopped operating after the \textit{Doorbell} task, our pilot only managed to solve two tasks, which eventually placed us third. The winning team, EyeRider, solved seven tasks, which is similar to our winning run in the qualification.
\bib


\section{Discussion}
\subsection{System Limitations and Real-world Readiness} \label{section:discussion}
Our proposed system successfully addresses the technical challenges of the Cybathlon competition, demonstrating promising performance across most tasks. While our evaluations confirm its effectiveness, deployment in a real-world use case still requires several adaptations.

The combination of depth cameras and RGB sensors, complemented by audio feedback and vibrations, has proven to be a highly effective combination. However, the current mechanical design remains at the prototype stage and requires further development to improve usability and wearability. A key limitation for usability is that the front plate and backpack are only connected by sensor cables, making it difficult for a blind user to put and remove the system without help from an additional person.

However, several components of our solution show potential for real-world application. The navigation system, combined with the vibration belt, is well-suited for deployment in everyday scenarios. Similarly, the color sorting function has been tested in various contexts and could be used as is. 

The doorbell and seat-finding solutions rely on specific assumptions about the environment, but could function effectively in a broader range of environments with some adaptions.

Other tasks, such as grocery item detection and touchscreen interaction, currently serve as proof-of-concept implementations. These solutions were taylored for Cybathlon and significant adaptations would be necessary to generalize them for diverse real-world settings.

To make our system more adaptable to real-world scenarios, we see great potential in using \ac{VLM} to enhance its capabilities. Integrating \ac{VLM}s could significantly improve scene understanding by incorporating information, e.g. about the number and types of objects present. This could then be complemented with our existing classical vision-based approaches. Additionally, this would eliminate the constraint of predefined object classes, allowing for deployment in a wide range of environments. However, it is essential to ensure that these enhancements can be implemented without causing significant delays to keep the system responsive to the user's movement.

\subsection{Lessons Learned} 
Throughout the development of our approach, we identified several key insights related to both the technical limitations and practical challenges of deploying a navigation and perception system for visual assistance.

In terms of sensing hardware, reliable depth sensing remains a significant challenge, particularly in the presence of black, narrow, or featureless objects. Monocular depth estimation methods still do not achieve accurate and robust metric depth at real-time frame rates, especially on low-power platforms. Recent stereo-based methods, such as FoundationStereo, demonstrate promising improvements. However, deploying such models on resource-constrained devices like the NVIDIA Jetson remains an open challenge.

We also observed that image quality has a substantial impact on the performance of perception systems, particularly those relying on learned approaches. While off-the-shelf neural networks performed well under ideal conditions (i.e. sharp, well-lit images), their robustness significantly deteriorated under common real-world degradations such as motion blur and low-light scenarios which are frequently encountered by a moving, blind person.

Another challenge we encountered during testing was the difficulty of conveying 3D spatial information to users through audio feedback. For example, if a target object appears in the bottom-right corner of the camera image and the goal is to bring the pilot closer to it, this can be achieved by either translating or rotating or a combination of both. Communicating such a motion through language is hard, as the human vocabulary for spatial directions (e.g., front, back, left, right, up, down) is overly coarse and primarily focused on translation, whereas users navigate in six degrees of freedom. Advanced stereo audio techniques, such as those used in cinematography, may offer more precise and intuitive spatial guidance in the future.

An additional challenge that emerged from the overall software architecture was that switching tasks within the system resulted in noticeable delays. This was primarily due to the reliance on development-stage code and the initialization times of neural networks. Replacing these components with more optimized implementations could significantly reduce the initialization times of individual subsystems and improve overall responsiveness.
\bib



\section{Conclusion}
This work presented Sight Guide, an assistive system developed for the \ac{VIS} at Cybathlon 2024. The system integrates data from a wearable multi-camera hardware setup in a modular, task-specific software architecture to enable visually impaired users to autonomously complete real-world inspired challenges. Our approach combines \ac{VIO}, 3D semantic mapping, object detection, \ac{OCR}, and interactive feedback through vibration and audio signals, to provide spatial and semantic information to the user. The device achieved a 95.7\% success rate in our controlled training environment and our results in the Cybathlon 2024, including a first-place qualification run validate the effectiveness of our approach. As our approach was optimized for the predefined competition environment, a generalization towards real-world environmental variability emerges as an immediate direction for future work. In particular, the adoption of more generalized perception models, such as vision-language models, could greatly enhance adaptability to unknown environments, though their performance may still be uncertain in the presence of blur, occlusion, or low-quality images commonly encountered in real-world deployments.

\bib



\section{Acknowledgements}
We would like to thank our pilot Lukas Hendry for his trust in the project, his patience throughout the testing process, his valuable feedback, and the significant time and effort he dedicated to supporting this work.

\bib



\bibliographystyle{IEEEtran} 
\bibliography{IEEEabrv,references.bib}

\end{document}